\documentclass[10pt,twocolumn,letterpaper]{article}

\usepackage[pagenumbers]{iccv} 

\definecolor{iccvblue}{rgb}{0.21,0.49,0.74}
\usepackage[pagebackref,breaklinks,colorlinks,allcolors=iccvblue]{hyperref}

\usepackage{array, booktabs}
\usepackage{multirow}
\usepackage[title]{appendix}

\usepackage{pifont}
\newcommand{\cmark}{\ding{51}}%
\newcommand{\xmark}{\ding{55}}%

\newcolumntype{L}[1]{>{\raggedright\let\newline\\\arraybackslash\hspace{0pt}}m{#1}}

\definecolor{figred}{RGB}{218,66,47}
\definecolor{figblue}{RGB}{52,127,179}

\def\paperID{8881} 
\def\confName{ICCV}
\def\confYear{2025}

\title{Fixing the RANSAC Stopping Criterion}

\author{Johannes Schönberger$^{1, 2}$\quad\quad Viktor Larsson$^3$\quad\quad Marc Pollefeys$^{1, 2}$\\[5pt]
$^1$ETH Zurich\quad\quad$^2$Microsoft\quad\quad$^3$Lund University
}

\begin{document}
\maketitle

\begin{abstract}

For several decades, RANSAC has been one of the most commonly used robust estimation algorithms for many problems in computer vision and related fields.
The main contribution of this paper lies in addressing
a long-standing error baked into virtually any system building upon the RANSAC algorithm.
Since its inception in 1981 by Fischler and Bolles, many variants of RANSAC have been proposed on top of the same original idea relying on the fact that random sampling has a high likelihood of generating a good hypothesis from minimal subsets of measurements.
An approximation to the sampling probability was originally derived by the paper in 1981 in support of adaptively stopping RANSAC and is, as such, used in the vast majority of today's RANSAC variants and implementations.
The impact of this approximation has since not been questioned or thoroughly studied by any of the later works.
As we theoretically derive and practically demonstrate in this paper, the approximation leads to severe undersampling and thus failure to find good models.
The discrepancy is especially pronounced in challenging scenarios with few inliers and high model complexity.
An implementation of computing the exact probability is surprisingly simple yet highly effective and has potentially drastic impact across a large range of computer vision systems.

\end{abstract}

\section{Introduction}

The problem of robust model estimation from noisy and outlier contaminated measurements is a prevalent challenge in many disciplines in engineering and science.
For many decades, the field has received great attention in the literature~\cite{huber1974robust,david1977exploratory,rousseeuw1984least,hough1962method,fischler1981random}.
Random Sample Consensus (RANSAC), originally proposed by Fischler and Bolles~\cite{fischler1981random} in 1981, has proven as one of the most influential works in computer vision.
RANSAC and its more modern variants are often the method of choice for robust estimation problems in computer vision, including but not limited to motion segmentation~\cite{torr1993outlier}, stereo~\cite{torr1998robust,matas2004robust}, geometric primitive fitting~\cite{sminchisescu2005incremental}, image mosaicing~\cite{szeliski1997creating}, multi-model fitting~\cite{isack2012energy,pham2014interacting,barath2019progressive}, camera localization~\cite{irschara2009structure,sattler2011fast}, and structure-from-motion~\cite{pollefeys2004visual,schonberger2016structure}.

\begin{figure}[t]
\centering
\includegraphics[width=\columnwidth]{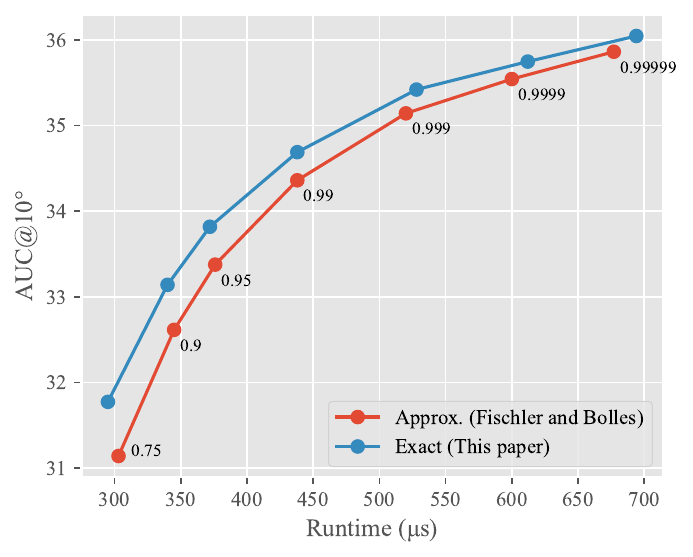}
\caption{
\textbf{Impact of approximation in RANSAC.}
The figure shows the accuracy (AUC@10$^\circ$, higher is better) against runtime for essential matrix estimation (see details in Section~\ref{sec:practical-impact}).
Each point on the curve corresponds to a target success probability $s$.
The original RANSAC paper (Fischler and Bolles~\cite{fischler1981random}) introduces an approximation for the all-inlier probability ({\bf\color{figred}approx}). 
This causes premature stopping for instances with few matches.
Using the correct ({\bf\color{figblue}exact}) probability avoids this and yields a consistent improvement.
Alternatively, it is possible to increase the hyperparameter $s$ such that the approximation reaches the same number of iterations as using the exact probability.
However, as the graph shows, this leads to slower convergence as easy instances are then unnecessarily oversampled.
In other words, one cannot simply adjust the target success probability when using the standard approximation.
}
\label{fig:teaser}
\end{figure}

The family of RANSAC algorithms follows a so-called \emph{hypothesize-and-verify} framework that iteratively hypothesizes models from randomly sampled measurements and verifies them against all observations to pick the hypothesis with maximum consensus.
The idea relies on the fact that random sampling has a high likelihood of generating a good hypothesis from minimal subsets of measurements.
Using uniform random sampling, the likelihood of generating a good hypothesis from a set of all-inliers depends on the expected outlier ratio and the minimal sample size imposed by the model complexity.
As the outlier ratio or the sample size increase, the number of required RANSAC iterations grows exponentially.
In practice, the underlying outlier ratio of the measurements is typically unknown \textit{a priori} and varies for each problem instance.
This makes it absolutely crucial to use an adaptive stopping criterion in RANSAC in order to successfully find good models within a limited compute budget.
The standard approach involves running RANSAC for a varying number of iterations, given an \textit{a priori} defined success probability of finding a good model and depending on the observed inlier ratio of the so far best sampled model.

Already in the original RANSAC paper by Fischler and Bolles, an adaptive stopping criterion is derived that depends on an approximate probability for sampling an all-inlier set of measurements.
To the best of our knowledge, the vast majority of follow-up literature and RANSAC implementations are based on this approximation.
As we theoretically and practically demonstrate in this paper, this approximation is overly optimistic and thus leads to an undersampling of model hypotheses.
Especially in challenging cases with few measurements and low inlier rates, the approximation error is large and often leads to failure in finding a good model.
Due to the error's non-linear relation \wrt the number of measurements and the inlier ratio, it cannot be accounted for by simply defining a higher success probability without then also oversampling (\ie, wasting compute) in the easy cases (\cf~Figure~\ref{fig:teaser}).

The main contribution of this paper lies in addressing a long-standing error baked into virtually any system that builds upon the RANSAC algorithm.
We provide a theoretical derivation of the error and present a principled fix to the problem.
The fix is surprisingly simple yet highly effective and has potentially drastic impact across a large range of computer vision systems.
In several synthetic and real experiments, we apply our proposed fix to different geometric computer vision problems to show the large potential.

The remainder of the paper is organized as follows.
First, we discuss related work in Section~\ref{sec:related-work} and introduce a derivation of the approximate and exact stopping criteria in Section~\ref{sec:stopping-criterion}.
Section~\ref{sec:theoretical-impact} then provides a theoretical analysis of the approximation error and its impact.
Next in Section~\ref{sec:practical-impact}, we evaluate the practical performance impact on various computer vision problems and in different benchmarks.
Finally, we conclude with a discussion of limitations in Section~\ref{sec:limitations}.

\section{Related Work}
\label{sec:related-work}

The RANSAC algorithm was originally proposed by Fischler and Bolles~\cite{fischler1981random} in 1981 as a framework for robust model estimation.
It has found widespread application in many computer vision problems~\cite{torr1993outlier,torr1998robust,matas2004robust,sminchisescu2005incremental,szeliski1997creating,isack2012energy,pham2014interacting,barath2019progressive,irschara2009structure,sattler2011fast,pollefeys2004visual,schonberger2016structure}.
Since then many improvements to the original idea have been proposed.
MLESAC~\cite{torr2000mlesac} combines the RANSAC algorithm with an maximum-likelihood based scoring instead of traditional inlier counting.
NAPSAC~\cite{torr2002napsac} introduces a locality-driven sampling strategy instead of uniform random sampling.
LO-RANSAC~\cite{chum2003loransac} proposes to locally optimize models based on a non-minimal refinement on the best inlier set.
PROSAC~\cite{chum2005matching} uses a progressive sampling approach using a ranking of measurements to more quickly find a good model.
Preemtive RANSAC~\cite{nister2005preemptive} proposes a preemption scheme and ARRSAC~\cite{raguram2008comparative} an adaptive sampling strategy for real-time applications.
R-RANSAC~\cite{matas2004randomized} uses randomized sampling and early model rejection based on the Sequential Probability Ratio Test (SPRT) for improved efficiency.
DEGENSAC~\cite{chum2005two} and QDEGSAC~\cite{frahm2006ransac} address the challenge of (quasi-)degenerate data while Cov-RANSAC~\cite{raguram2009exploiting} presents a principled approach to leveraging measurement uncertainties.
GroupSAC~\cite{ni2009groupsac} clusters the data by similarity to improve efficiency in low inlier scenarios.
USAC~\cite{raguram2013usac} combines the best of several RANSAC variants into a unified framework.
In recent years, many more RANSAC variants (GC-RANSAC~\cite{barath2018graph}, P-NAPSAC~\cite{barath2019progressive}, MAGSAC~\cite{barath2018magsac,barath2020magsac++}, DSAC~\cite{brachmann2017dsac}, VSAC~\cite{ivashechkin2021vsac}, Space-Partitioning RANSAC~\cite{barath2022space}) have been proposed to improve the efficiency, robustness, and other aspects.
Common to all the above RANSAC variants, including the original one, is the use of an adaptive stopping criterion using an approximation of the probability of sampling an all-inlier model.
The fact that the probability is approximated is a known fact, however, has not been studied or questioned anywhere in the literature.
As we will show in this paper, the approximation produces significant errors in challenging scenarios with low inlier rates and when dealing with complex models.


Apart from the many proposed approaches in the literature, there are also a myriad of existing implementations of RANSAC and its variants available as open source.
This includes standalone releases of approaches like USAC, MAGSAC, GC-RANSAC, \etc as well as implementations of RANSAC in frameworks such as OpenCV~\cite{opencv_library} or scikit-image~\cite{van2014scikit} and end-to-end systems like COLMAP~\cite{schonberger2016structure}, OpenMVG~\cite{moulon2016openmvg}, or ORB-SLAM~\cite{ORBSLAM3_TRO}.
To the best of our knowledge, the vast majority and, in particular, the most popular open source implementations of RANSAC implement the discussed approximation and can thus benefit from our proposed fix to the stopping criterion.
We provide a comprehensive list of references to the relevant source code in the supplementary material.
In practice, our proposed fix will most often require only a few lines of change in the source code of available implementations.

\begin{figure*}[h]
\centering
\includegraphics[width=\textwidth]{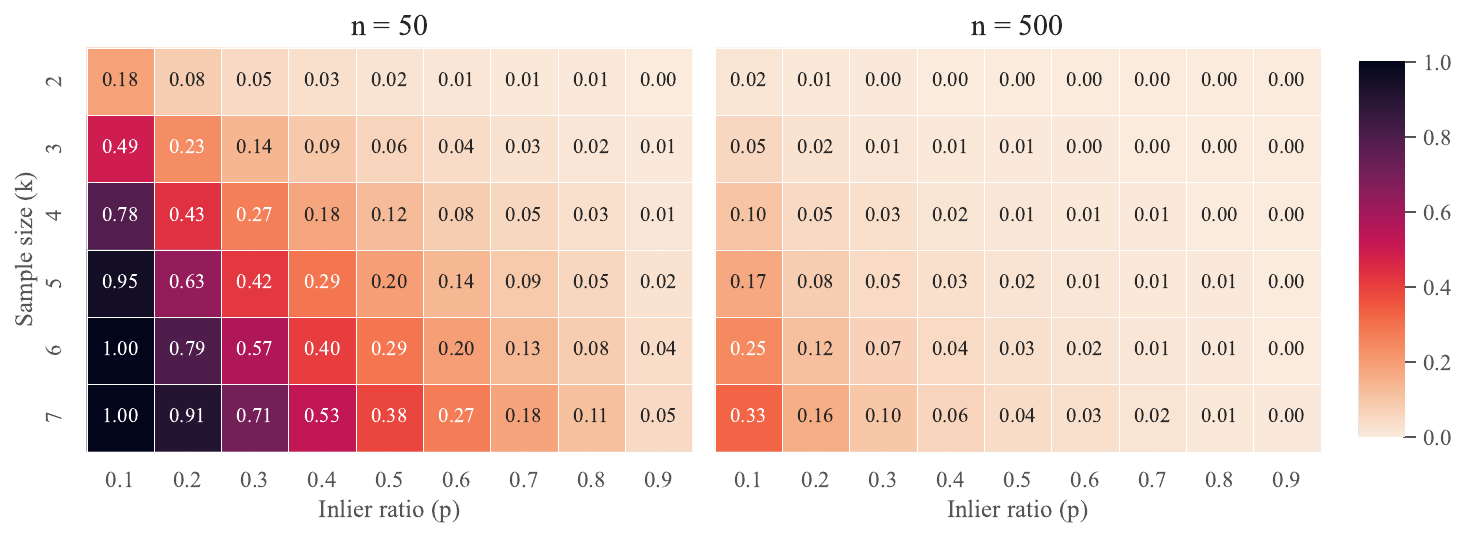}
\caption{\textbf{Relative error $\epsilon = \tfrac{P_a-P_e}{P_a}$ in all-inlier probabilities.} The error introduced by the approximation as a function of sample size $k$ and inlier ratio $p$. With few measurements (\textit{Left}, $n=50$), the approximation is much worse. With more measurements (\textit{Right}, $n=500$), the difference between the approximate and exact probability only appears for low inlier ratios and high sample sizes.}
\label{fig:relative_approx_error}
\end{figure*}

\section{Stopping Criterion in RANSAC}
\label{sec:stopping-criterion}

Let $s$ denote the desired probability that our sampling is successful (i.e.~we draw at least one all-inlier sample) and let $P$ denote the probability of drawing an all-inlier sample.
Then the number of trials $N$ should satisfy
\begin{equation}
    (1-P)^N \leq 1-s  
\end{equation}
Taking the logarithm and noting that $(1-P)<1$ we get
\begin{equation}
    N \ge \frac{\log(1-s)}{\log(1-P)}
\end{equation}
The original RANSAC paper~\cite{fischler1981random} suggested to use
\begin{equation} \label{eq:approx}
    P_a = p^k
\end{equation}
for the all-inlier probability $P$,
where $p$ is the inlier ratio and $k$ the number of sampled measurements. 
This has since become the standard approach for computing the required number of iterations in RANSAC.

However, this only provides the approximate probability $P_a$, as drawing an inlier measurement for our sample changes the inlier ratio when sampling another measurement in the same iteration.
Using uniform random sampling, the exact probability $P_e$ can be computed as the ratio between the number of all-inlier samples and the number of possible samples, \ie
\begin{equation}
    P_e = \frac{{p n \choose k}}{{n \choose k}}
\end{equation}
or equivalently formulated as
\begin{equation}
    P_e = \prod_{i=0}^{k-1} \frac{p n - i}{n - i} \quad\text{if}~ pn \ge k ~\text{and 0 otherwise,}
\end{equation}
where $n$ is the total number of measurements and $p n$ the total number of inliers.
From a practical standpoint, the latter formula can be implemented without special care using fixed-point arithmetic.
For all but the most extreme scenarios, 32 bits for integer and floating point numbers provide sufficient numerical precision.
It is important to note that the calculation of the exact probability has marginal computational overhead as compared to the approximation.
In practice, the difference should not be measurable due to hypothesis generation and verification requiring orders of magnitude more compute for any real problem.

\begin{figure*}[h]
\centering
\includegraphics[height=0.34\textwidth]{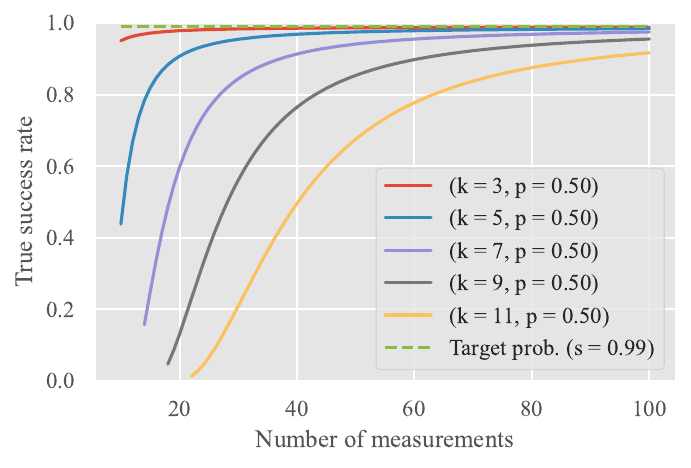}
\includegraphics[height=0.34\textwidth]{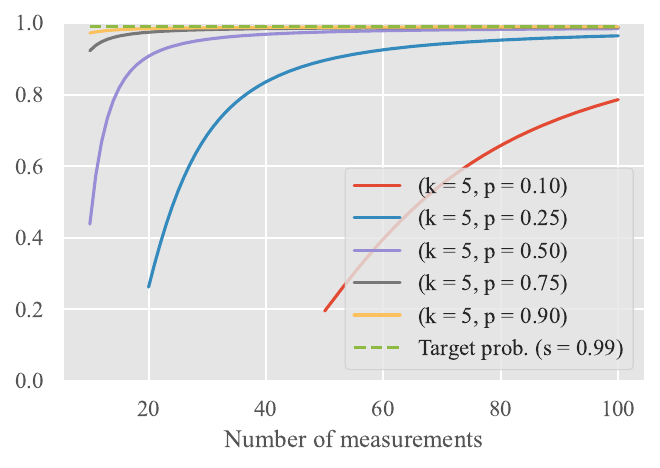}
\caption{\textbf{True success rate $s_{true}$ as a function of number of measurements.} The graphs show the actual success rate attained when using the classic stopping criteria in RANSAC. For instances with fewer measurements, the approximate probability leads to early exits. }
\label{fig:true_success_rate}
\end{figure*}

\begin{figure}[h]
\centering
\includegraphics[width=\columnwidth]{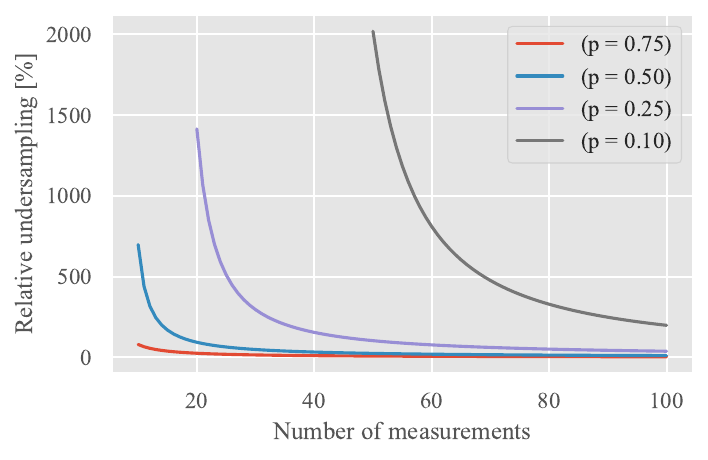}
\caption{\textbf{Amount of undersampling.} The graph shows the relative difference of required iterations for the exact versus the approximate stopping criterion as $N_{e/a} = \tfrac{N_e - N_a}{N_a}$ to reach a target success probability $s = 0.99$ with a sample size $k = 5$. The approximation leads to severe undersampling in low inlier scenarios.}
\label{fig:undersampling}
\end{figure}

For problems with inlier-only measurements ($p = 1$) or when using 1-point RANSAC~\cite{scaramuzza20111,schonberger2017vote} ($k = 1$), the approximation is exact, \ie,~$P_a = P_e$.
In the general case, it is easy to see that the approximation strictly overestimates the exact probability
\begin{equation}
    P_e = \prod_{i=0}^{k-1} \frac{pn - i}{n - i} < \prod_{i=0}^{k-1} \frac{p(n - i)}{n - i} = p^k = P_a \enspace ,
\end{equation}
since $pi < i$ for positive $i$.
Here, we can also easily see why the approximation seems to be often working well in practice.
At the limit, where the number of measurements is large compared to the sample size ($n \gg k$) or the inlier probability is high ($p \approx 1$), we have 
\begin{equation}
\frac{P_e}{P_a} = \prod_{i=0}^{k-1} \frac{pn - i}{pn - pi} \approx 1 \enspace .
\end{equation}
It is important to highlight that this formula does not only apply to the original RANSAC but also more modern variants like PROSAC, ARRSAC, Preemptive RANSAC, R-RANSAC, SPRT, GC-RANSAC, VSAC, MAGSAC, MAGSAC++, Space Partitioned RANSAC, \etc.

In this paper, we will show that, when operating in scenarios with few inliers or complex models requiring a large sample size, the classic RANSAC stopping criterion often leads to catastrophic early stopping failing to achieve the desired success probability.
First, in Section~\ref{sec:theoretical-impact}, we investigate the impact of the approximation error on the number of iterations and resulting success-rate. Next, in Section~\ref{sec:practical-impact}, we experimentally show that using the correct probability makes a difference in practice.

\section{Theoretical Analysis}
\label{sec:theoretical-impact}

In this section, we provide a theoretical study of the error introduced by using the approximate probability $P_a$ instead of the exact sampling probability $P_e$.
First, we compare the relative difference between the probabilities, \ie,
\begin{equation}
    \epsilon = \frac{P_a - P_e}{P_a} \enspace,
\end{equation}
which is always non-negative as $P_a \ge P_s$, as seen above.
The approximate probability $P_a$ only depends on the inlier ratio $p$ and sample size $k$, while $P_e$ additionally depends on the number of measurements $n$.

Figure~\ref{fig:relative_approx_error} visualizes the relative error $\epsilon$ for different parameter choices.
In instances with few measurements $(n=50)$, the approximation error grows significantly as the sample size increases or the inlier probability drops.
For larger problems ($n=500$), the approximation works well in all but the most extreme scenarios.

The RANSAC stopping criterion is derived by defining a target probability that our sampling is successful.
A natural question is now, if we run the number of iterations computed from the approximate probability, what is the actual success rate that we attain. 
This can be computed as
\begin{equation}
    s_{true} = 1 - {(1 - P_e)}^\frac{\log(1-s)}{\log(1-P_a)}
\end{equation}
where $s$ is the target probability.
Note that using the exact probability $P_e$ instead of $P_a$ above gives $s_{true} = s$.
Figure~\ref{fig:true_success_rate} shows how the true success rate $s_{true}$ varies with the number of measurements for different configurations (sample size $k$ and inlier probability $p$).
Similarly, Figure~\ref{fig:undersampling} shows the relative percentage by which the approximation undersamples the exact stopping criterion.
Notice that the curves cut off at $n = \tfrac{k}{p}$ measurements, because no all-inlier sample exists for fewer measurements.
In practice, unless the sampled measurements provide more constraints than minimally necessary, RANSAC needs at minimum $k + 1$ inliers to verify that a sampled hypothesis is a good one using the additional inlier.
In both figures, we can see that the classic stopping criterion severely undersamples, thus leading to many failed instances to find a good model.


\begin{table*}
\begin{center}
\begin{tabular}{c c c c c c c c c c r} \toprule
Inlier ratio & \multicolumn{3}{c}{AUC@1} & \multicolumn{3}{c}{AUC@2} & \multicolumn{3}{c}{AUC@3} & $\Delta$ Time \\
\cmidrule(lr){2-4} \cmidrule(lr){5-7} \cmidrule(lr){8-10}
(p) & Approx. & Exact & $\Delta$  & Approx. & Exact & $\Delta$ & Approx. & Exact & $\Delta$ & ($N_{e/a}$) \\\midrule
0.2 & 36.144 & 36.709 & \color{gray}{+1.56}\% & 57.057 & 57.782 & \color{gray}{+1.27}\% & 67.783 & 68.497 & \color{gray}{+1.05}\% & +7.2\% \\
0.3 & 42.238 & 42.696 & \color{gray}{+1.08}\% & 65.013 & 65.483 & \color{gray}{+0.72}\% & 75.777 & 76.163 & \color{gray}{+0.51}\% & +4.8\% \\
0.4 & 44.542 & 44.833 & \color{gray}{+0.65}\% & 67.426 & 67.732 & \color{gray}{+0.45}\% & 77.778 & 78.022 & \color{gray}{+0.31}\% & +1.6\% \\
0.5 & 45.744 & 45.899 & \color{gray}{+0.34}\% & 68.639 & 68.792 & \color{gray}{+0.22}\% & 78.730 & 78.850 & \color{gray}{+0.15}\% & +0.8\% \\
\bottomrule
\end{tabular}
\end{center}
\caption{\textbf{Synthetic 2D line fitting experiment.} End-to-end AUC (higher is better) with relative improvements $\Delta$ and runtime ($\Delta N$) metrics computed over 100K simulated problem instances with each $n = 50$ measurements and $s = 0.99$ success probability. 
}
\label{tab:synthetic-line}
\end{table*}

\begin{table*}
\begin{center}
\begin{tabular}{c c c c c c c c c c r} \toprule
Inlier ratio & \multicolumn{3}{c}{AUC@1} & \multicolumn{3}{c}{AUC@2} & \multicolumn{3}{c}{AUC@3} & $\Delta$ Time \\
\cmidrule(lr){2-4} \cmidrule(lr){5-7} \cmidrule(lr){8-10}
(p) & Approx. & Exact & $\Delta$  & Approx. & Exact & $\Delta$ & Approx. & Exact & $\Delta$ & ($N_{e/a}$) \\\midrule
0.2 & 14.033 & 14.600 & \color{gray}{+4.04}\% & 24.414 & 25.714 & \color{gray}{+5.32}\% & 31.154 & 33.053 & \color{gray}{+6.10}\% & +49.4\% \\
0.3 & 32.002 & 33.202 & \color{gray}{+3.75}\% & 53.570 & 56.222 & \color{gray}{+4.95}\% & 65.583 & 68.146 & \color{gray}{+3.91}\% & +21.4\% \\
0.4 & 34.992 & 36.035 & \color{gray}{+2.98}\% & 57.757 & 59.833 & \color{gray}{+3.59}\% & 70.009 & 71.832 & \color{gray}{+2.60}\% & +15.4\% \\
0.5 & 36.256 & 37.068 & \color{gray}{+2.24}\% & 60.236 & 61.422 & \color{gray}{+1.97}\% & 72.475 & 73.486 & \color{gray}{+1.39}\% & +7.7\% \\
\bottomrule
\end{tabular}
\end{center}
\caption{\textbf{Synthetic 2D ellipse fitting experiment.} End-to-end AUC (higher is better) with relative improvements $\Delta$ and runtime ($\Delta N$) metrics computed over 100K simulated problem instances with each $n = 100$ measurements and $s = 0.99$ success probability.
}
\label{tab:synthetic-ellipse}
\end{table*}

\section{Practical Impact} \label{sec:practical-impact}

In this section, we perform synthetic and real experiments on different computer vision problems.
In contrast to the theoretical analysis in the previous section, we measure the impact of the approximation error using end-to-end metrics relevant to the target application.

\subsection{Geometric Primitive Fitting}

In this first experiment, we study the impact of the approximation error based on the problem of geometric 2D line ($k = 2$) and 2D ellipse ($k = 5$) fitting using synthetically generated measurements $\mathbf{x}$.
To this end, we synthesize problem instances with random line segments and ellipses within the box $[-100, 100] \times [-100, 100]$.
From the synthetic parametric models, we sample ground-truth points $\mathbf{\tilde{x}} \in \mathbb{R}^2$ with varying ratio of inliers and added isotropic Gaussian noise $\mathcal{N}(0, \mathbf{\Sigma_x})$ generated s.t.~$\vert \mathbf{\Sigma_x} \vert \in [0.5, 2]$.
The remaining outliers follow a uniform random distribution within the sampling box.
We use an implementation of the original RANSAC algorithm~\cite{fischler1981random} and compare the resulting best models obtained with the standard approximate and our proposed exact stopping criterion.
The residuals between measurements and hypotheses are computed as $\mathbf{r} = \mathbf{x} - \mathbf{\tilde{x}}$ with measurements being labeled as inliers if its whitened distance $X = \sqrt{\mathbf{r}^T \mathbf{\Sigma_x^{-1}} \mathbf{r}} \leq 3$.
The end-to-end error metric is reported over a large number of problem instances as the area-under-the-curve ($AUC@X$) at different confidence intervals.
Please note that the relative performance difference at the largest confidence interval is a good measure of the number of catastrophic estimation failures due to undersampling.
For each problem instance, we compute one error sample per ground-truth point as the shortest Euclidean distance to the estimated best model.
In addition, we also report the relative runtime increase as a function of the number of performed RANSAC iterations as $N_{e/a} = \tfrac{N_e - N_a}{N_a}$ until the approximate and exact stopping criteria are satisfied.

Tables~\ref{tab:synthetic-line}~and~\ref{tab:synthetic-ellipse} summarize the experimental results.
We observe that the theoretical analysis in the previous section translates into practice, where the exact stopping criterion consistently recovers better models across all settings.
Note that the practical improvement is not fully consistent with the theory, as our analysis ignores secondary effects (\eg measurement noise) impacting the sampling probability.
As expected, the relative gap between the exact and approximate criteria increases with lower inlier ratios.
Similarly, the gap is larger for the more complex ellipse model ($k = 5$) as compared to the less complex line model ($k = 2$).
This is achieved by running RANSAC for the required number of iterations in the difficult low inlier cases while only adding little runtime overhead in the easier cases with more inliers.
For example, in the case of the 2D line experiments, the exact criterion only leads to 0.8\% more iterations for an inlier ratio of 50\% while 7.2\% more iterations are performed in the difficult 20\% inlier scenario.
The results for AUC@3 at the largest confidence interval show that a significant portion of models catastrophically fail due to the approximation.
For example, the exact implementation recovers 6.1\% more ellipse models for an inlier ratio of $p=0.2$.

\subsection{Camera Pose Estimation}

In this section, we perform various experiments on relative and absolute camera pose estimation (homography, essential matrix, fundamental matrix, Perspective-n-Point) using PoseLib's~\cite{PoseLib} benchmarking suite.
The suite evaluates the performance of PoseLib's state-of-the-art minimal pose solvers embedded inside a standard LO-RANSAC~\cite{chum2003loransac} robust estimator on multiple datasets (see Table~\ref{tbl:datasets}).
The only adjustment to PoseLib is the change of the adaptive stopping criterion.
For the experiments, we do not set a minimum or maximum number of iterations \textit{a priori} but control the runtime only with the approximate and exact stopping criteria.
The results are reported as the area-under-the-curve ($AUC@X$) at different accuracy thresholds $X$ together with the average runtime.
Again, the relative performance difference at the largest error thresholds provides a good measure of the number of catastrophic estimation failures due to undersampling.

The results of the experiments are summarized in Tables~\ref{tab:relative-pose} and~\ref{tab:absolute-pose}.
For each problem, we vary the number of measurements by uniform random sub-sampling as well as the target success probability.
Consistently, across all configurations, the exact stopping criterion leads to a better result than the approximation.
As expected from the theoretical analysis, the runtime and performance gap increases significantly for the more challenging cases with fewer inliers and measurements while there is no measurable difference for the easier cases with many measurements.
For example, the case of homography estimation with $n=20$ and $s = 0.95$, one obtains a consistent 10\% relative performance improvement across all accuracy thresholds.
At the largest error thresholds, we observe that the approximation leads to a significant portion of catastrophic estimation failures due to undersampling.
Results at the smaller accuracy thresholds show that, where undersampling does not lead to catastrophic failures, the exact sampling can also boost the results.

Further, when comparing the different estimation problems, we observe the same trend as before: the impact of the approximation varies with model complexity ($k = 3$ for Perspective-n-Point, $k = 4$ for homography, $k = 5$ for essential matrix, $k = 7$ for fundamental matrix).
Notice that the inlier ratio of the benchmark data for the homography evaluation is significantly lower than the ones for essential and fundamental matrix estimation, thus leading to a relatively larger accuracy improvement.

When comparing the performance and runtime impact of the approximate vs.~exact stopping criteria with the different evaluated target success probabilities $s$, notice that the approximation error cannot be simply mitigated by increasing $s$, as it will also increase the runtime for both the easy and the hard cases.
This is further illustrated in Figure~\ref{fig:teaser} which shows this trade-off between accuracy and runtime as the success probability $s$ varies.  
The data for the figure is from essential matrix estimation on on ScanNet1500 with $n=50$ SIFT matches.

\begin{table}
\resizebox{\columnwidth}{!}{
\begin{tabular}{l l l} \toprule
& Dataset & \\
\midrule
\multirow{3}{*}{\rotatebox{90}{\bf E / F}}
& MegaDepth-1500~\cite{li2018megadepth,sun2021loftr} & \footnotesize SIFT \cite{lowe2004distinctive}, SP+SG/LG \cite{detone2018superpoint,sarlin2020superglue,lindenberger2023lightglue} \\
& ScanNet-1500 \cite{dai2017scannet,sarlin2020superglue} &  \footnotesize SIFT \cite{lowe2004distinctive}, SP+SG\cite{detone2018superpoint,sarlin2020superglue}\\
& IMC 2021 PT \cite{imc2021data} &\footnotesize SIFT\cite{lowe2004distinctive} \\ \midrule
\multirow{1}{*}{\bf \rotatebox{90}{H}} & \multirow{1}{*}{HEB\cite{barath2023large}} (Alamo, NYC Library) & \footnotesize SIFT\cite{lowe2004distinctive} \\  \midrule
\multirow{3}{*}{\rotatebox{90}{\bf P3P}} & ETH3D \cite{schops2017multi,dusmanu2020multi} &\footnotesize SIFT\cite{lowe2004distinctive} \\
& 7Scenes \cite{shotton2013scene} (Heads, Stairs)  &\footnotesize  HLoc\cite{sarlin2019coarse} (NetVLAD\cite{arandjelovic2016netvlad}+SP+SG) \\
& Cambridge Landmarks \cite{kendall2015posenet} &\footnotesize  HLoc\cite{sarlin2019coarse}  (NetVLAD\cite{arandjelovic2016netvlad}+SP+SG)\\ \bottomrule
\end{tabular}
}
\caption{\textbf{Datasets.} For the experimental evaluation of camera pose estimation, we use instances from multiple standard benchmark datasets with different keypoint detectors and matchers.}
\label{tbl:datasets}
\end{table}

\begin{table*}
\begin{center}
\resizebox{\textwidth}{!}{
\begin{tabular}{c c c c c c c c c c c c c} \toprule
\multicolumn{2}{c}{Parameters} & \multicolumn{3}{c}{AUC@5} & \multicolumn{3}{c}{AUC@10} & \multicolumn{3}{c}{AUC@20} & \multicolumn{2}{c}{Average~Runtime} \\
\cmidrule(lr){1-2} \cmidrule(lr){3-5} \cmidrule(lr){6-8} \cmidrule(lr){9-11} \cmidrule(lr){12-13}
n & s & Approx. & Exact & $\Delta$ & Approx. & Exact & $\Delta$ & Approx. & Exact & $\Delta$ & Approx. & Exact \\\midrule
\multicolumn{13}{c}{Homography} \\\midrule
\multirow{ 2}{*}{20} & 0.95 & 2.99 & 3.31 & \color{gray}{+10.70}\% & 5.87 & 6.48 & \color{gray}{+10.39}\% & 9.94 & 10.91 & \color{gray}{+9.76}\% & 0.2ms & 0.7ms \\
 & 0.99 & 3.25 & 3.51 & \color{gray}{+8.00}\% & 6.23 & 6.74 & \color{gray}{+8.19}\% & 10.53 & 11.27 & \color{gray}{+7.03}\% & 0.3ms & 0.7ms \\
\multirow{ 2}{*}{100} & 0.95 & 10.03 & 10.03 & \color{gray}{+0.00}\% & 17.32 & 17.33 & \color{gray}{+0.06}\% & 25.19 & 25.20 & \color{gray}{+0.04}\% & 1.0ms & 1.0ms \\
 & 0.99 & 10.10 & 10.13 & \color{gray}{+0.30}\% & 17.51 & 17.53 & \color{gray}{+0.11}\% & 25.39 & 25.39 & \color{gray}{+0.00}\% & 1.0ms & 1.0ms \\
\multirow{ 2}{*}{200} & 0.95 & 11.69 & 11.70 & \color{gray}{+0.09}\% & 19.79 & 19.80 & \color{gray}{+0.05}\% & 28.18 & 28.18 & \color{gray}{+0.00}\% & 1.2ms & 1.2ms \\
 & 0.99 & 11.72 & 11.72 & \color{gray}{+0.00}\% & 19.82 & 19.82 & \color{gray}{+0.00}\% & 28.14 & 28.14 & \color{gray}{+0.00}\% & 1.2ms & 1.2ms \\
\multirow{ 2}{*}{1000} & 0.95 & 58.49 & 58.49 & \color{gray}{+0.00}\% & 76.44 & 76.44 & \color{gray}{+0.00}\% & 88.21 & 88.21 & \color{gray}{+0.00}\% & 3.9ms & 4.0ms \\
 & 0.99 & 58.78 & 58.78 & \color{gray}{+0.00}\% & 76.55 & 76.55 & \color{gray}{+0.00}\% & 88.27 & 88.27 & \color{gray}{+0.00}\% & 4.3ms & 4.3ms \\
\midrule
\multicolumn{13}{c}{Essential matrix} \\\midrule
\multirow{ 2}{*}{20} & 0.95 & 25.27 & 25.87 & \color{gray}{+2.37}\% & 38.63 & 39.42 & \color{gray}{+2.05}\% & 52.15 & 53.02 & \color{gray}{+1.67}\% & 0.4ms & 0.8ms \\
 & 0.99 & 26.08 & 26.63 & \color{gray}{+2.11}\% & 39.66 & 40.36 & \color{gray}{+1.77}\% & 53.26 & 54.01 & \color{gray}{+1.41}\% & 0.6ms & 1.4ms  \\
\multirow{ 2}{*}{100} & 0.95 & 45.78 & 45.93 & \color{gray}{+0.33}\% & 60.81 & 61.00 & \color{gray}{+0.31}\% & 73.11 & 73.29 & \color{gray}{+0.25}\% & 2.0ms & 2.1ms    \\
 & 0.99 & 46.77 & 46.89 & \color{gray}{+0.26}\% & 61.86 & 62.01 & \color{gray}{+0.24}\% & 74.07 & 74.21 & \color{gray}{+0.19}\% & 2.1ms & 2.2ms    \\
\multirow{ 2}{*}{200} & 0.95 & 49.89 & 49.95 & \color{gray}{+0.12}\% & 64.39 & 64.47 & \color{gray}{+0.12}\% & 76.00 & 76.06 & \color{gray}{+0.08}\% & 2.2ms & 2.2ms    \\
 & 0.99 & 50.73 & 50.76 & \color{gray}{+0.06}\% & 65.38 & 65.41 & \color{gray}{+0.05}\% & 76.91 & 76.94 & \color{gray}{+0.04}\% & 2.3ms & 2.3ms    \\
\multirow{ 2}{*}{1000} & 0.95 & 37.88 & 37.87 & \color{gray}{+-0.03}\% & 51.83 & 51.83 & \color{gray}{+0.00}\% & 63.18 & 63.18 & \color{gray}{+0.00}\% & 1.8ms & 1.8ms    \\
 & 0.99 & 38.19 & 38.19 & \color{gray}{+0.00}\% & 52.48 & 52.48 & \color{gray}{+0.00}\% & 63.97 & 63.96 & \color{gray}{+-0.02}\% & 2.0ms & 2.0ms    \\
\midrule
\multicolumn{13}{c}{Fundamental matrix} \\\midrule
\multirow{ 2}{*}{20} & 0.95 & 8.59 & 9.02 & \color{gray}{+5.01}\% & 15.98 & 16.67 & \color{gray}{+4.32}\% & 26.50 & 27.44 & \color{gray}{+3.55}\% & 0.2ms & 0.5ms \\
 & 0.99 & 8.98 & 9.36 & \color{gray}{+4.23}\% & 16.60 & 17.22 & \color{gray}{+3.73}\% & 27.35 & 28.17 & \color{gray}{+3.00}\% & 0.3ms & 0.7ms \\
\multirow{ 2}{*}{100} & 0.95 & 24.62 & 24.79 & \color{gray}{+0.69}\% & 37.32 & 37.56 & \color{gray}{+0.64}\% & 50.97 & 51.24 & \color{gray}{+0.53}\% & 1.4ms & 1.4ms   \\
 & 0.99 & 25.38 & 25.54 & \color{gray}{+0.63}\% & 38.35 & 38.57 & \color{gray}{+0.57}\% & 52.11 & 52.36 & \color{gray}{+0.48}\% & 1.5ms & 1.5ms   \\
\multirow{ 2}{*}{200} & 0.95 & 30.09 & 30.16 & \color{gray}{+0.23}\% & 43.54 & 43.63 & \color{gray}{+0.21}\% & 57.08 & 57.17 & \color{gray}{+0.16}\% & 1.9ms & 1.9ms   \\
 & 0.99 & 30.86 & 30.93 & \color{gray}{+0.23}\% & 44.54 & 44.62 & \color{gray}{+0.18}\% & 58.22 & 58.31 & \color{gray}{+0.15}\% & 3.1ms & 2.0ms   \\
\multirow{ 2}{*}{1000} & 0.95 & 26.37 & 26.37 & \color{gray}{+0.00}\% & 40.42 & 40.42 & \color{gray}{+0.00}\% & 54.85 & 54.85 & \color{gray}{+0.00}\% & 2.2ms & 2.2ms   \\
 & 0.99 & 26.57 & 26.58 & \color{gray}{+0.04}\% & 40.44 & 40.46 & \color{gray}{+0.05}\% & 54.37 & 54.38 & \color{gray}{+0.02}\% & 2.3ms & 2.3ms \\
\bottomrule
\end{tabular}
}
\end{center}
\caption{\textbf{Relative camera pose estimation.} Two-view homography ($k = 4$), essential matrix ($k = 5$), fundamental matrix estimation ($k = 7$) results with varying number of measurements $n$ and target success probability $s$ reported as end-to-end AUC (higher is better) with relative improvements $\Delta$ and runtime metrics.
\vspace{10pt}
}
\label{tab:relative-pose}
\end{table*}

\begin{table*}
\begin{center}
\resizebox{\textwidth}{!}{
\begin{tabular}{c c c c c c c c c c c c c} \toprule
\multicolumn{2}{c}{Parameters} & \multicolumn{3}{c}{AUC@5} & \multicolumn{3}{c}{AUC@10} & \multicolumn{3}{c}{AUC@20} & \multicolumn{2}{c}{Average~Runtime} \\
\cmidrule(lr){1-2} \cmidrule(lr){3-5} \cmidrule(lr){6-8} \cmidrule(lr){9-11} \cmidrule(lr){12-13}
n & s & Approx. & Exact & $\Delta$ & Approx. & Exact & $\Delta$ & Approx. & Exact & $\Delta$ & Approx. & Exact \\\midrule
\multirow{ 2}{*}{20} & 0.95 & 26.90 & 27.10 & \color{gray}{+0.74}\% & 76.97 & 77.16 & \color{gray}{+0.25}\% & 93.62 & 93.71 & \color{gray}{+0.10}\% & 51us & 52us \\
 & 0.99 & 27.30 & 27.43 & \color{gray}{+0.48}\% & 77.19 & 77.37 & \color{gray}{+0.23}\% & 93.79 & 93.91 & \color{gray}{+0.13}\% & 57us & 58us \\
\multirow{ 2}{*}{100} & 0.95 & 43.77 & 43.80 & \color{gray}{+0.07}\% & 88.10 & 88.13 & \color{gray}{+0.03}\% & 97.41 & 97.42 & \color{gray}{+0.01}\% & 0.2ms & 0.2ms\\
 & 0.99 & 43.84 & 43.84 & \color{gray}{+0.00}\% & 88.31 & 88.32 & \color{gray}{+0.01}\% & 97.49 & 97.49 & \color{gray}{+0.00}\% & 0.2ms & 0.2ms \\
\multirow{ 2}{*}{200} & 0.95 & 47.96 & 47.96 & \color{gray}{+0.00}\% & 89.76 & 89.77 & \color{gray}{+0.01}\% & 97.82 & 97.82 & \color{gray}{+0.00}\% & 0.3ms & 0.3ms \\
 & 0.99 & 47.99 & 48.00 & \color{gray}{+0.02}\% & 89.88 & 89.88 & \color{gray}{+0.00}\% & 97.85 & 97.85 & \color{gray}{+0.00}\% & 0.4ms & 0.3ms \\
\multirow{ 2}{*}{1000} & 0.95 & 50.48 & 50.48 & \color{gray}{+0.00}\% & 91.54 & 91.54 & \color{gray}{+0.00}\% & 98.24 & 98.24 & \color{gray}{+0.00}\% & 1.5ms & 1.5ms \\
 & 0.99 & 50.54 & 50.54 & \color{gray}{+0.00}\% & 91.56 & 91.56 & \color{gray}{+0.00}\% & 98.25 & 98.25 & \color{gray}{+0.00}\% & 1.7ms & 1.7ms \\
\bottomrule
\end{tabular}
}
\end{center}
\caption{\textbf{Absolute camera pose estimation.} Perspective-n-Point estimation ($k = 3$) results with varying number of measurements $n$ and target success probability $s$ reported as end-to-end AUC (higher is better) with relative improvements $\Delta$ and runtime metrics.
}
\label{tab:absolute-pose}
\end{table*}

\section{Discussion and Limitations}
\label{sec:limitations}

This paper addresses an important error that has been neglected in virtually all existing RANSAC literature and implementations.
The approximation always leads to an undersampling and, therefore, the exact solution comes with a runtime increase.
However, by implementation of the exact stopping criterion, one can achieve the actually desired target success probability and thus consistently recover significantly more good models across all problems and scenarios.
The performance gap is especially pronounced in the challenging cases.

In our theoretical analysis, we ignore commonly known other factors, such as measurement noise, minimal solver stability, adaptive sampling, \etc, which can lead to a discrepancy between the desired and actually achieved target success probability.
The theory in proposed adaptive sampling strategies like PROSAC, ARRSAC, Preemptive RANSAC, R-RANSAC, SPRT, \etc are based on the same approximation as the original RANSAC paper.
As such, they can directly benefit from an implementation of the exact stopping criterion, but it is beyond the scope of this paper to analyze the practical impact of our proposed fix on the exhaustive list of RANSAC variants.
Furthermore, the impact of inlier measurement noise has been studied and various mitigations have been proposed in the literature~(\eg~\cite{torr2002napsac,raguram2009exploiting,raguram2011recon,barath2020magsac++}).
Measurement noise typically leads to a lower observed inlier probability $p$ and thus to oversampling with both the approximate and exact stopping criteria.
While some of these factors have been discussed before, not all of them are well known or addressed in a principled manner.
Combining the different factors impacting the achieved target success probability into a unified framework is an important problem to be addressed.

\section{Conclusion}

In this paper, we are the first to thoroughly study and address a problem present in virtually all existing RANSAC literature and implementations.
We provide a theoretical analysis of the problem with a derived exact solution to RANSAC's adaptive stopping criterion.
In experiments on both synthetic and real data, we show a significant and consistent practical negative impact of the classic approximation.
The impact is especially pronounced for problems with complex models and in the challenging cases with few inliers.
The implementation of the exact stopping criterion presented in this paper is surprisingly simple yet highly effective.
For known publicly available open-source implementations, we plan to submit code changes to raise awareness and benefit the overall community.

\begin{appendices}
\section{Open Source Implementations}

In Table~\ref{tab:ransac_list}, we provide a (non-exhaustive) list of references to the most important open source projects with RANSAC implementations.
For each project, we provide a URL to the specific location implementing the stopping criterion.
The URLs refer to the respective latest commits in the projects' main branches.
In only two cases, were we able to find implementations using the correct inlier probabilities.
All other projects will benefit from changing the stopping criterion.
Notice that the RANSAC implementations in these projects are consumed by many other projects, which thus transitively benefit from our proposed fix.
We omit references to repositories with RANSAC implementations based on an a priori fixed number of iterations.

\newcommand{\approxprob}{{\color{red}\xmark}}
\newcommand{\exactprob}{{\color{green}\cmark}}
\newcommand{\maybeapproxprob}{{\color{orange}\xmark}}

\begin{table*}
\begin{tabular}{L{\dimexpr0.12\linewidth-2\tabcolsep}
                c
                L{\dimexpr0.82\linewidth-2\tabcolsep}} 
                \toprule
\bf Project & \bf Correct? & \bf Repository link \\ \toprule
\textbf{\footnotesize OpenCV} & \approxprob &{\tiny\url{https://github.com/opencv/opencv/blob/1b0d58a554da4cb6ed613260811445ad9020a430/modules/calib3d/src/usac/termination.cpp#L44-L49}} \\ \midrule
 \textbf{\footnotesize scikit-image} & \approxprob& {\tiny\url{https://github.com/scikit-image/scikit-image/blob/0e97db25f3638b923be9b470fc5232403b4ffa95/skimage/measure/fit.py#L640-L667}} \\ \midrule
    \textbf{\footnotesize scikit-learn} & \approxprob& {\tiny\url{https://github.com/scikit-learn/scikit-learn/blob/6bf2061f76ba0977c1f7ffb9ddc48db794e5c7ec/sklearn/linear_model/_ransac.py#L48-L79}} \\  \midrule
    \textbf{\footnotesize kornia} & \approxprob& {\tiny\url{https://github.com/kornia/kornia/blob/b2edb53ec7db87fedfb78a395d7b096a33e09cd4/kornia/geometry/ransac.py#L98-L106}} \\  \midrule
     \textbf{\footnotesize RansacLib} & \approxprob&{\tiny\url{https://github.com/tsattler/RansacLib/blob/b952083fcf1b2aea001bf178e010e540ff372889/RansacLib/utils.h#L110-L140}} \\ \midrule
     \textbf{\footnotesize MAGSAC} & \approxprob&{\tiny\url{https://github.com/danini/magsac/blob/db121cf3b84431a2e14846c5a387941a70af6730/src/pymagsac/include/magsac.h#L938-L940}} \\
    & &{\tiny\url{https://github.com/danini/magsac/blob/db121cf3b84431a2e14846c5a387941a70af6730/src/pymagsac/include/magsac.h#L1122-L1129}} \\ \midrule
    \textbf{\footnotesize Progressive-X} & \approxprob& {\tiny\url{https://github.com/danini/progressive-x/blob/691a367ce68f7ba829047ab8fc49f98112ffe39d/src/pyprogressivex/include/progressive_x.h#L491-L513}} \\ \midrule
    \textbf{\footnotesize GC-RANSAC} & \approxprob& {\tiny\url{https://github.com/danini/graph-cut-ransac/blob/15e8b1b723a772ff0439489e65b419265df325b9/src/pygcransac/include/estimators/sample_consensus_estimator.h#L216-L242}} \\ \midrule
    \textbf{\footnotesize VSAC} & \approxprob&{\tiny\url{https://github.com/ivashmak/vsac/blob/229e7b9e28ad6818155e577acbb31bd5c9a88065/src/termination.cpp#L26-L36}} \\
    && {\tiny\url{https://github.com/ivashmak/vsac/blob/229e7b9e28ad6818155e577acbb31bd5c9a88065/src/termination.cpp#L80-L116}}
    \\&& {\tiny\url{https://github.com/ivashmak/vsac/blob/229e7b9e28ad6818155e577acbb31bd5c9a88065/src/termination.cpp#L176-L188}}
    \\
    && {\tiny\url{https://github.com/ivashmak/vsac/blob/229e7b9e28ad6818155e577acbb31bd5c9a88065/src/termination.cpp#L292-L341}} \\ \midrule
    \textbf{\footnotesize COLMAP}& \approxprob &{\tiny\url{https://github.com/colmap/colmap/blob/d065cea09317b2c1cbb9fad66376464119a3ee07/src/colmap/optim/ransac.h#L155-L179}} \\ \midrule
    \textbf{\footnotesize Theia} & \approxprob&{\tiny\url{https://github.com/sweeneychris/TheiaSfM/blob/d2112f15aa69a53dda68fe6c4bd4b0f1e2fe915b/src/theia/solvers/sample_consensus_estimator.h#L214-L243}} \\ \midrule
    \textbf{\footnotesize OpenMVG} & \approxprob&{\tiny\url{https://github.com/openMVG/openMVG/blob/e4f0c3dba12d07da2eb3b186629b02f5985421c9/src/openMVG/robust_estimation/robust_ransac_tools.hpp#L49-L57}} \\ \midrule
    \textbf{\footnotesize Meshroom} & \approxprob&{\tiny\url{https://github.com/alicevision/AliceVision/blob/b44ea5bf46aa03cc8d736398bb32ecdcdd3c0193/src/aliceVision/robustEstimation/ransacTools.hpp#L29-L32}} \\ \midrule
    \textbf{\footnotesize OpenSfM} & \approxprob&{\tiny\url{https://github.com/mapillary/OpenSfM/blob/ef872b2399cf1cc036d91e950f210a7be33c2745/opensfm/src/robust/robust_estimator.h#L20-L35}} \\ \midrule
    \textbf{\footnotesize ORB-SLAM3} & \approxprob& {\tiny\url{https://github.com/UZ-SLAMLab/ORB_SLAM3/blob/4452a3c4ab75b1cde34e5505a36ec3f9edcdc4c4/src/MLPnPsolver.cpp#L247-L255}} \\ \midrule
    \textbf{\footnotesize PoseLib} & \approxprob&{\tiny\url{https://github.com/PoseLib/PoseLib/blob/c86d70d75b6b9ec2652e4c327b3692b059f7400d/PoseLib/robust/ransac_impl.h#L106-L116}} \\ \midrule
    \textbf{\footnotesize OpenGV} & \approxprob&{\tiny\url{https://github.com/laurentkneip/opengv/blob/91f4b19c73450833a40e463ad3648aae80b3a7f3/include/opengv/sac/implementation/Ransac.hpp#L94-L113}}\\
    && 
    {\tiny\url{https://github.com/laurentkneip/opengv/blob/91f4b19c73450833a40e463ad3648aae80b3a7f3/include/opengv/sac/implementation/MultiRansac.hpp#L93-L123}}
    \\ \midrule
    \textbf{\footnotesize Open3D} & \approxprob& {\tiny\url{https://github.com/isl-org/Open3D/blob/e86fcb38ddb5cb5e3d77c53cab234ddd687fbcfd/cpp/open3d/pipelines/registration/Registration.cpp#L238-L242}} \\ \midrule
    \textbf{\footnotesize PCL} & \approxprob& {\tiny\url{https://github.com/PointCloudLibrary/pcl/blob/0932486c52a2cf4f0821e25d5ea2d5767fff8381/sample_consensus/include/pcl/sample_consensus/impl/ransac.hpp#L161-L176}} \\ \midrule
    \textbf{\footnotesize CGAL} & \maybeapproxprob& {\tiny\url{https://github.com/CGAL/cgal/blob/4b1c13011fe894a1167ce3b7901303a00cdf6f68/Shape_detection/include/CGAL/Shape_detection/Efficient_RANSAC/Efficient_RANSAC.h#L1066-L1071}} \\ \midrule
     \textbf{\footnotesize pyransac} & \exactprob & {\tiny\url{https://github.com/ducha-aiki/pydegensac/blob/030a44dc66eb5906a6898cc4b23cc5544b7244fe/src/pydegensac/degensac/rtools.c#L202-L225}}
    \\ \midrule
\textbf{\footnotesize USAC} & \exactprob & {\tiny\url{https://github.com/cr333/usac-cmake/blob/0fd1934b696088b273ac7fe4dea400a07e45475c/src/estimators/USAC.h#L922-L995}} \\
\bottomrule
\end{tabular}
\caption{Non-exhaustive list of open-source RANSAC implementations. The CGAL implementation is not using the standard RANSAC approximation but is most likely still incorrect due to a bug.}
\label{tab:ransac_list}
\end{table*}

\end{appendices}

{
    \small
    \bibliographystyle{ieeenat_fullname}
    \bibliography{main}
}

\end{document}